\documentclass{article}
\usepackage{ijcai19}

\usepackage{times}
\usepackage{soul}
\usepackage{url}
\usepackage[utf8]{inputenc}
\usepackage[small]{caption}
\usepackage{graphicx}
\usepackage{amsmath}
\usepackage{booktabs}
\usepackage{threeparttable}
\usepackage{amsmath}
\usepackage{multirow}
\usepackage{color}
\usepackage[colorlinks,
            linkcolor=black,
            anchorcolor=black,
            citecolor=black,
            urlcolor=blue
            ]{hyperref}
\title{StructADMM: A Systematic, High-Efficiency Framework of Structured Weight Pruning for DNNs}

\author{Tianyun Zhang$^{1*}$, Shaokai Ye$^{2*}$, Kaiqi Zhang$^1$, Xiaolong Ma$^4$, Ning Liu$^4$, Linfeng Zhang$^3$, Jian Tang$^1$,Kaisheng Ma$^3$, Xue Lin$^4$, Makan Fardad$^1$ \& Yanzhi Wang$^4$\\
\affiliations
1. Syracuse University 2. Sensetime 3. Tsinghua University \\ 4. Northeastern University  $^*$Equal Contribution\\}

\begin{document}

\maketitle

\begin{abstract}
Weight pruning methods of DNNs have been demonstrated to achieve a good model pruning rate without loss of accuracy, thereby alleviating the significant computation/storage requirements of large-scale DNNs. Structured weight pruning methods have been proposed to overcome the limitation of 
irregular network structure
and demonstrated actual GPU acceleration. %measurements.
However, in prior work the pruning rate (degree of sparsity) and GPU acceleration are limited (to less than 50\%) when accuracy needs to be maintained.
In this work, we overcome these limitations by proposing a unified, systematic framework of structured weight pruning for DNNs. It is a framework that can be used to induce different types of structured sparsity, such as filter-wise, channel-wise, and shape-wise sparsity, as well non-structured sparsity. The proposed framework incorporates stochastic gradient descent with ADMM, and can be understood as a dynamic regularization method in which the regularization target is analytically updated in each iteration.
% %
% A significant improvement in structured weight pruning ratio is achieved without loss of accuracy, along with fast convergence rate. 
%
% With a small sparsity degree of 33.3\% on the convolutional layers, we achieve 1.64\% accuracy enhancement for the AlexNet (CaffeNet) model.
% %
% This is obtained  by mitigation of overfitting.
%
Without loss of accuracy on the AlexNet model, we achieve 2.58$\times$ and 3.65$\times$ average measured speedup on two GPUs, clearly outperforming the prior work. The average speedups reach 3.15$\times$ and 8.52$\times$ when allowing a moderate accuracy loss of 2\%. In this case the model compression for convolutional layers is 15.0$\times$, corresponding to 11.93$\times$ measured CPU speedup. Our experiments on ResNet model and on other datasets like UCF101 and CIFAR-10 demonstrate the consistently higher performance of our framework. 
\end{abstract}

\section{Introduction}
Deep neural networks (DNNs) utilize multiple functional layers cascaded together to extract features at multiple levels of abstraction \cite{krizhevsky2012imagenet,simonyan2014}, and are thus both computationally and storage intensive. As a result, many studies on DNN model compression are underway, including weight pruning \cite{han2015,DeepCompression,dai2017,wen2016learning,yang2016}, low-rank approximation \cite{denil2013predicting,denton2014exploiting,wen2017coordinating}, low displacement rank approximation (structured matrices) \cite{cheng2015exploration,sindhwani2015structured,zhao2017theoretical}, etc.
Weight pruning can achieve a high model pruning rate without loss of accuracy.
A pioneering work \cite{han2015,DeepCompression} adopts an iterative weight pruning heuristic and results in a sparse neural network structure. It can achieve 9$\times$ weight reduction with no accuracy loss on AlexNet \cite{krizhevsky2012imagenet}. This weight pruning method has been extended in \cite{dai2017,yang2016,guo2016dynamic,dong2017learning} to either use more sophisticated algorithms to achieve a higher weight pruning rate, or to obtain a fine-grained trade-off between a higher pruning rate and a lower accuracy degradation.

Despite the promising results, these general weight pruning methods often produce non-structured and irregular connectivity in DNNs. This leads to degradation in the degree of parallelism and actual performance in GPU and hardware platforms. Moreover, the weight pruning rate is mainly achieved through compressing the fully-connected (FC) layers \cite{han2015,DeepCompression,guo2016dynamic}, which are less computationally intensive compared with convolutional (CONV) layers and are becoming less important in state-of-the-art DNNs such as ResNet \cite{he2016deep}.
To address these limitations, recent work \cite{wen2016learning,he2017channel} have proposed to learn structured sparsity, including sparsity at the levels of filters, channels, filter shapes, layer depth, etc. These works focus on CONV layers and actual GPU speedup is reported as a result of structured sparsity \cite{wen2016learning}. However, these structured weight pruning methods are based on regularization techniques and are still quite heuristic \cite{wen2016learning,he2017channel}. The weight pruning rate and GPU acceleration are both quite limited. For example, the average weight pruning rate on CONV layers of AlexNet is only 1.5$\times$ without any accuracy loss, corresponding to 33.3\% sparsity. 

In this work, we overcome this limitation by proposing \emph{a unified, systematic framework of structured weight pruning for DNNs}, named StructADMM, based on the powerful optimization tool Alternating Direction Method of Multipliers (ADMM) \cite{boyd2011,hong2017linear} shown to perform well on combinatorial constraints. It is a unified framework for different types of structured sparsity such as filter-wise, channel-wise, and shape-wise sparsity, as well as non-structured sparsity. It is a systematic framework of dynamic ADMM regularization and masked mapping and retraining steps, guaranteeing solution feasibility (satisfying all constraints) and providing high solution quality. It achieves a significant improvement in weight pruning rate under the same accuracy, along with fast convergence rate.
 In the context of deep learning, the StructADMM framework can be understood as a smart and dynamic regularization technique in which the regularization target is analytically updated in each iteration.

We conduct extensive experiments using ImageNet data set on two GPUs (NVIDIA 1080Ti and Jetson TX2), and Intel i7-6700K CPU.
Without accuracy loss on the AlexNet model (with over 2\% accuracy enhancement in baseline accuracy), we achieve 2.58$\times$ and 3.65$\times$ average measured speedup on two GPUs, which clearly outperforms the GPU acceleration of 49\% reported in SSL \cite{wen2016learning}.
The speedups reach 3.15$\times$ and 8.52$\times$ when allowing a moderate accuracy loss of 2\%. In this case the model compression for CONV layers is 15.0$\times$, corresponding to 11.93$\times$ CPU speedup. Our experiments on ResNet model on ImageNet dataset and on other datasets like UCF101, MNIST, and CIFAR-10 demonstrate the consistently higher performance (accuracy, structured pruning rate) of our framework, compared with prior work and with the \emph{proximal gradient descent} (PGD) method.

We share our codes and models at link \textcolor{blue}{\url{http://bit.ly/2M0V7DO}}, and supplemental material at \textcolor{blue}{\url{http://bit.ly/2IBiNBk}}.

\section{Related work}

\textbf{\emph{General, non-structured weight pruning.}} The pioneering work by Han \emph{et al.} \cite{han2015,DeepCompression} achieved 9$\times$ reduction in the number of parameters in AlexNet and 13$\times$ in VGG-16. However, most reduction is achieved in FC layers, and 2.7$\times$ reduction achieved in CONV layers will not lead to an overall acceleration in GPU \cite{wen2016learning}. 
Extensions of iterative weight pruning, such as \cite{guo2016dynamic} (dynamic network surgery), \cite{dai2017} (NeST) and \cite{mao2017exploring}, use more delicate algorithms such as selective weight growing and pruning. But the weight pruning rates on CONV layers are still limited, e.g., 3.1$\times$ in \cite{guo2016dynamic}, 3.23$\times$ in \cite{dai2017}, and 4.16$\times$ in \cite{mao2017exploring} for AlexNet with no accuracy degradation. This level of non-structured weight pruning cannot guarantee GPU acceleration. In fact, our StructADMM framework can achieve 16.1$\times$ non-structured weight pruning in CONV layers of AlexNet without accuracy degradation, however, only minor GPU acceleration is actually observed.

\textbf{\emph{Structured weight pruning.}}
To overcome the limitation in non-structured, irregular weight pruning, SSL \cite{wen2016learning} proposes to learn structured sparsity at the levels of filters, channels, filter shapes, layer depth, etc. This work is one of the first with actually measured GPU accelerations. This is because CONV layers after structured pruning will transform to a full matrix multiplication in GPU (with reduced matrix size). 
However, the weight pruning rate and GPU acceleration are both limited. The average weight pruning rate on CONV layers of AlexNet is only 1.5$\times$ without accuracy loss. The reported GPU acceleration is 49\%. Besides, the recent work \cite{he2017channel} achieves 2$\times$ channel pruning with 1\% accuracy degradation on VGGNet.

\section{Structured Pruning Formulation as Optimization Problems}

Consider an $N$-layer DNN in which the first $M$ layers are CONV layers and the rest are FC layers. The weights and biases of the $i$-th layer are respectively denoted by ${\bf{W}}_{i}$ and ${\bf{b}}_{i}$, and the loss function associated with the DNN is denoted by $f \big( \{{\bf{W}}_{i}\}_{i=1}^N, \{{\bf{b}}_{i} \}_{i=1}^N \big)$; see \cite{zhang2018systematic}. In this paper, $\{{\bf{W}}_{i}\}_{i=1}^N$ and $\{{\bf{b}}_{i} \}_{i=1}^N$ respectively characterize the collection of weights and biases from layer $1$ to layer $N$.

In this paper, our objective is to implement structured pruning on the DNN. In the following discussion we focus on the CONV layers because they have the highest computation requirements.
More specifically, we minimize the loss function subject to specific structured sparsity constraints on the weights in the CONV layers, i.e., 
\begin{equation}
\label{opt0}
\begin{aligned}
& \underset{ \{{\bf{W}}_{i}\},\{{\bf{b}}_{i} \}}{\text{minimize}}
& & f \big( \{{\bf{W}}_{i}\}_{i=1}^N, \{{\bf{b}}_{i} \}_{i=1}^N \big),
\\ & \text{subject to}
& & {\bf{W}}_{i}\in {\bf{S}}_{i}, \; i = 1, \ldots, M,
\end{aligned}
\end{equation}
where ${\bf{S}}_{i}$ is the set of ${\bf{W}}_{i}$ with a specific ``structure". Next we introduce constraint sets corresponding to different types of structured sparsity. Non-structured, irregular sparsity constraints (i.e., pruning) are also included in the framework. The suitability for GPU acceleration is discussed for different types of sparsity, and we finally introduce the proper combination of structured sparsities to facilitate GPU accelerations.

\begin{figure}[b]
\centering
\includegraphics[width=\linewidth]{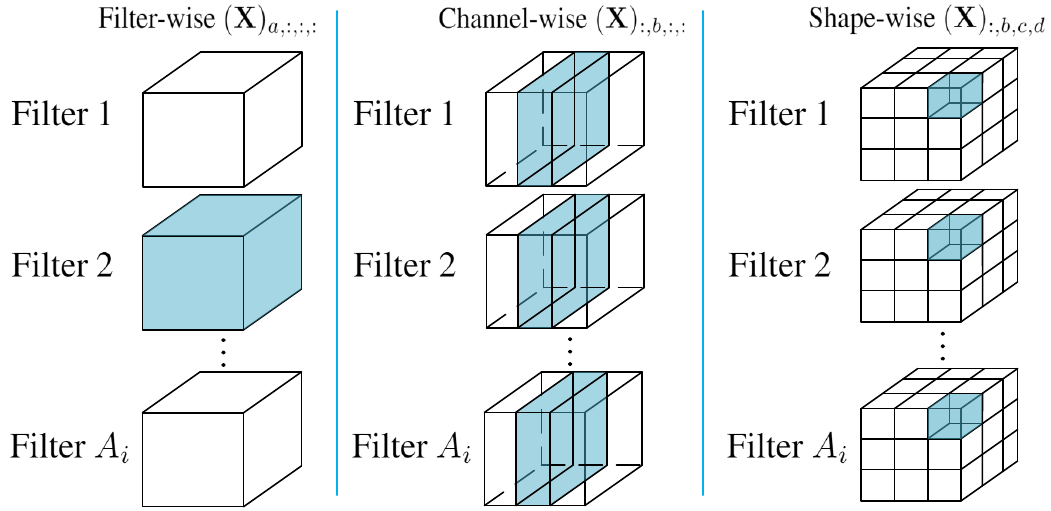}
\caption{Illustration of filter-wise, channel-wise and shape-wise structured sparsities from left to right.}
\label{fig:example}
\end{figure}

\begin{figure*} 
\centering
\includegraphics[width=0.6\linewidth]{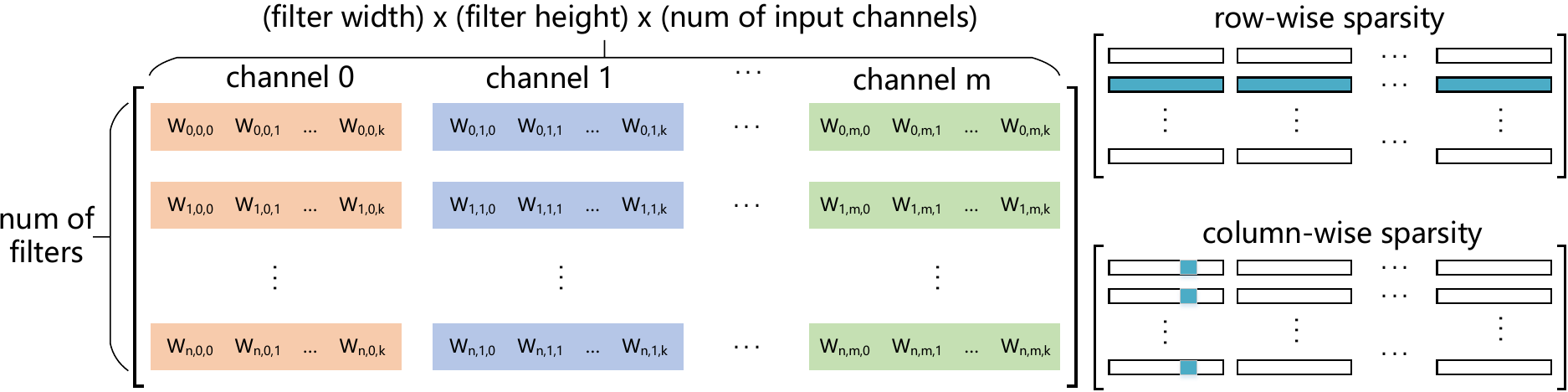}
\caption{ Illustration of 2D weight matrix for GEMM (left) and row-wise and column-wise sparsity (right)}
\label{fig:example}
\end{figure*}

The collection of weights in the $i$-th CONV layer is a four-dimensional tensor, i.e., ${\bf{W}}_{i} \in R^{A_i \times B_i \times C_i \times D_i}$, where $A_i, B_i, C_i$, and $D_i$ are respectively the number of filters, the number of channels in a filter, the height of the filter, and the width of the filter, in layer $i$.
In what follows,
if $\bf{X}$ denotes the weight tensor in a specific layer, let $({\bf{X}})_{a,:,:,:}$ denote the $a$-th filter in $\bf{X}$, $({\bf{X}})_{:,b,:,:}$ denote the $b$-th channel, and $({\bf{X}})_{:,b,c,d}$ denote the collection of weights located at position $(:,b,c,d)$ in every filter of $\bf{X}$,
as illustrated in Figure 1.

{\emph{\textbf{Filter-wise structured sparsity}}}:
When we train a DNN with sparsity at the filter level, the constraint on the weights in the $i$-th CONV layer is given by ${\bf{W}}_{i}\in {\bf{S}}_{i}
:=
\{{\bf{X}}\mid \text{the number of nonzero filters in}$ ${\bf{X}} ~\text{is less than or equal to}$ $\alpha_i \}.$ Here, nonzero filter means that the filter contains some nonzero weight.

{\emph{\textbf{Channel-wise structured sparsity}}}: When we train a DNN with sparsity at the channel level, the constraint on the weights in the $i$-th CONV layer is given by
${\bf{W}}_{i} \in {\bf{S}}_{i} :=  \{{\bf{X}}\mid \text{the number of nonzero channels in}$ ${\bf{X}}~ \text{is less than or equal to}$ $\beta_{i}  \}.$
Here, we call the $b$-th channel nonzero if $({\bf{X}})_{:,b,:,:}$ contains some nonzero element.

{\emph{\textbf{Filter shape-wise structured sparsity}}}:
When we train a DNN with sparsity at the filter shape level, the constraint on the weights in the $i$-th CONV layer is given by
${\bf{W}}_{i}\in {\bf{S}}_{i}
:=
 \{{\bf{X}}\mid \text{the number of nonzero vectors in}$ $\{{\bf{X}}_{:,b,c,d}\}_{b,c,d=1}^{B_i,C_i,D_i}$ $ \text{is less than or equal to}~ \theta_i  \}.$
 
{\emph{\textbf{Non-structured, irregular weight sparsity}}}: When we train a DNN with non-structured weight sparsity, the constraint on the weights in $i$-th CONV layer is
$
{\bf{W}}_{i} \in {\bf{S}}_{i} := \{{\bf{X}}\mid \text{number of nonzero elements in}$ ${\bf{X}}~ \text{is less than or equal to}$ $\gamma_{i}  \}.$

{\emph{\textbf{Combination of structured sparsities to facilitate GPU acceleration}}}:
Convolutional computations in DNNs are commonly transformed to matrix multiplications by converting weight tensors and feature map tensors to matrices \cite{chetlur2014cudnn}, named \emph{general matrix multiplication} or GEMM. Filter-wise sparsity corresponds to row pruning, whereas channel-wise and filter shape-wise sparsities correspond to column pruning in GEMM. The GEMM matrix maintains a full matrix with number of rows/columns reduced, thereby enabling GPU acceleration. In the results we will use \emph{row pruning} to represent the results of filter-wise sparsity in GEMM, and use \emph{column pruning} to represent the results of channel-wise and filter shape-wise sparsities.

\section{The Unified StructADMM Framework}

In problem (\ref{opt0}) the constraint is combinatorial. 
As a result, this problem cannot be solved directly by stochastic gradient descent methods like original DNN training.
However, the property that ${\bf{W}}_{i}$ satisfies certain combinatorial ``structures" is compatible with ADMM which is shown to be an effective method to deal with such clustering-like constraints \cite{hong2016convergence,liu2018zeroth}

Despite the compatibility, there is still challenge in the direction application of ADMM due to the non-convexity in objective function. To overcome this challenge, we propose StructADMM, a systematic framework of dynamic ADMM regularization and masked mapping and retraining steps. We can guarantee solution feasibility (satisfying all constraints) and provide high solution quality through this integration.

\emph{\textbf{ADMM Regularization Step}}: Corresponding to every set ${\bf{S}}_{i}$, $i = 1, \ldots, M$ we define the indicator function
$g_{i}({\bf{W}}_{i})=
\begin{cases}
 0 & \text { if } {\bf{W}}_{i}\in {\bf{S}}_{i}, \\ 
 +\infty & \text { otherwise.}
\end{cases}$
Furthermore, we incorporate auxilliary variables ${\bf{Z}}_{i}$, $i = 1, \ldots, M$.
The original problem (\ref{opt0}) is then equivalent to 
\begin{equation}
\label{admm_form}
\begin{aligned}
& \underset{ \{{\bf{W}}_{i}\},\{{\bf{b}}_{i} \}}{\text{minimize}}
& & f \big( \{{\bf{W}}_{i} \}_{i=1}^N, \{{\bf{b}}_{i} \}_{i=1}^N \big)+\sum_{i=1}^{M} g_{i}({\bf{Z}}_{i}),
\\ & \text{subject to}
& & {\bf{W}}_{i}={\bf{Z}}_{i}, \; i = 1, \ldots, M.
\end{aligned}
\end{equation}

Through formation of the augmented Lagrangian \cite{boyd2011}, the ADMM regularization decomposes problem (\ref{admm_form}) into two subproblems, and solves them iteratively until convergence\footnote{The details of ADMM are presented in \cite{boyd2011,zhang2018systematic}. We omit the details due to space limitation.}. The first subproblem is
\begin{equation}
\label{subproblem_1}
 \underset{ \{{\bf{W}}_{i}\},\{{\bf{b}}_{i} \}}{\text{minimize}}
\ \ \ f \big( \{{\bf{W}}_{i} \}_{i=1}^N, \{{\bf{b}}_{i} \}_{i=1}^N \big)+\sum_{i=1}^{M} \frac{\rho_{i}}{2}  \| {\bf{W}}_{i}-{\bf{Z}}_{i}^{k}+{\bf{U}}_{i}^{k} \|_{F}^{2}, \\
\end{equation}
where ${\bf{U}}_{i}^{k}:={\bf{U}}_{i}^{k-1}+{\bf{W}}_{i}^{k}-{\bf{Z}}_{i}^{k}$.
The first term in the objective function of (\ref{subproblem_1}) is the differentiable loss function of the DNN, and the second term is a quadratic regularization term of the ${\bf{W}}_{i}$'s, which is differentiable and convex. As a result (\ref{subproblem_1}) can be solved by stochastic gradient descent as original DNN training. Although we cannot guarantee the global optimality, it is due to the non-convexity of the DNN loss function rather than the quadratic term enrolled by our method. Please note that this subproblem and solution are the same for all types of structured sparsities.

On the other hand, the second subproblem is given by
\begin{equation}
 \underset{ \{{\bf{Z}}_{i} \}}{\text{minimize}}
\ \ \ \sum_{i=1}^{M} g_{i}({\bf{Z}}_{i})+\sum_{i=1}^{M} \frac{\rho_{i}}{2} \| {\bf{W}}_{i}^{k+1}-{\bf{Z}}_{i}+{\bf{U}}_{i}^{k} \|_{F}^{2}. \\
\end{equation}
Note that $g_{i}(\cdot)$ is the indicator function of ${\bf{S}}_{i}$, thus this subproblem can be solved analytically and optimally \cite{boyd2011}. For $i = 1, \ldots, M$, the optimal solution is the Euclidean projection of ${\bf{W}}_{i}^{k+1}+{\bf{U}}_{i}^{k}$ onto ${\bf{S}}_{i}$.
The set ${\bf{S}}_{i}$ is different when we apply different types of structured sparsity, and the Euclidean projections will be described next.

\emph{\textbf{Solving the second subproblem for different structured sparsities}}: For \textbf{filter-wise structured sparsity} constraints, we first calculate $O_{a}=\| ({\bf{W}}_{i}^{k+1} +{\bf{U}}_{i}^{k})_{a,:,:,:} \|_F^2$ for $a = 1, \ldots, A_i$, where $\| \cdot \|_F$ denotes the Frobenius norm. We then keep $\alpha_i$ elements in $({\bf{W}}_{i}^{k+1} +{\bf{U}}_{i}^{k})_{a,:,:,:}$ corresponding to the $\alpha_i$ largest values in $\{O_a\}_{a = 1}^{A_i}$ and set the rest to zero.

For \textbf{channel-wise structured sparsity}, we first calculate $O_{b}=\| ({\bf{W}}_{i}^{k+1} +{\bf{U}}_{i}^{k})_{:,b,:,:} \|_F^2$ for $b= 1, \ldots, B_i$. We then keep $\beta_{i}$ elements in $({\bf{W}}_{i}^{k+1} +{\bf{U}}_{i}^{k})_{:,b,:,:}$ corresponding to the $\beta_{i}$ largest values in $\{O_b\}_{b=1}^{B_i}$ and set the rest to zero.

For \textbf{shape-wise structured sparsity}, we first calculate $O_{b,c,d}=\| ({\bf{W}}_{i}^{k+1} + {\bf{U}}_{i}^{k})_{:,b,c,d} \|_F^2$ for $b= 1, \ldots, B_i$, $c= 1, \ldots, C_i$ and $d= 1, \ldots, D_i$. We then keep $\theta_i$ elements in $({\bf{W}}_{i}^{k+1} + {\bf{U}}_{i}^{k})_{:,b,c,d}$ corresponding to the $\theta_i$ largest values in $\{O_{b,c,d}\}_{b,c,d=1}^{B_i,C_i,D_i}$ and set the rest to zero.

For \textbf{non-structured weight sparsity}, we keep $\gamma_i$ elements in ${\bf{W}}_{i}^{k+1}+{\bf{U}}_{i}^{k}$ with largest magnitudes and set the rest to zero \cite{boyd2011,zhang2018systematic}.

\emph{\textbf{Increasing $\rho$ in ADMM regularization}}: The $\rho_{i}$ values are the most critical hyperparameter in ADMM regularization. We start from smaller $\rho_{i}$ values, say $\rho_{1} = \dots = \rho_{N} = 1.5\times 10^{-3}$, and gradually increase with ADMM iterations. This coincides with the theory of ADMM convergence \cite{hong2016convergence,liu2018zeroth}. It in general takes 8 - 12 ADMM iterations for convergence, corresponding to 100 - 150 epochs in PyTorch, a moderate increase compared with the original DNN training.

\emph{\textbf{Masked mapping and retraining}}: After ADMM regularization, we obtain intermediate ${\bf{W}}_{i}$ solutions. In this step, we first perform the said Euclidean projection (mapping) to guarantee that the structured pruning constraints are satisfied. Next, we mask the zero weights and retrain the DNN with non-zero weights using training sets (while keeping the masked weights 0). In this way test accuracy (solution quality) can be partially restored.

\subsection{Explanation of StructADMM in the deep learning context}
The proposed StructADMM is different from the conventional utilization of ADMM, i.e., to accelerate the convergence of an originally convex problem \cite{boyd2011,hong2017linear}. Rather, we propose to integrate the ADMM framework with stochastic gradient descent. Aside from recent mathematical optimization results \cite{hong2016convergence,liu2018zeroth} illustrating the advantage of ADMM with combinatorial constraints, the advantage of the proposed StructADMM framework can be explained in the deep learning context as described below.

The proposed StructADMM (\ref{subproblem_1}) can be understood as a smart, dynamic $L_2$ regularization method, in which the regularization target $ {\bf{Z}}_{i}^{k}-{\bf{U}}_{i}^{k}$ will change judiciously and analytically in each iteration. On the other hand, conventional regularization methods (based on $L_1$, $L_2$ norms or their combinations) use a fixed regularization target, and the penalty is applied on all the weights. This will inevitably cause accuracy degradation. In contrast, StructADMM will not penalize the remaining, nonzero weights as long as there are enough weights pruned to zero. For an illustration, Figure 3 demonstrates the comparison results (accuracy loss vs. pruned parameters) on the non-structured pruning of AlexNet model. Methods to compare include iterative pruning, two regularizations with retraining, and Projected Gradient Descent (PGD). We use the same hyperparameters in StructADMM as baseline regularizations and PGD for fairness. It is clear that StructADMM outperforms the others, while regularization-based methods even result in lower performance than \cite{han2015} because they will penalize all the weights.

\begin{figure}[ht]
\centering
\includegraphics[width=1\linewidth]{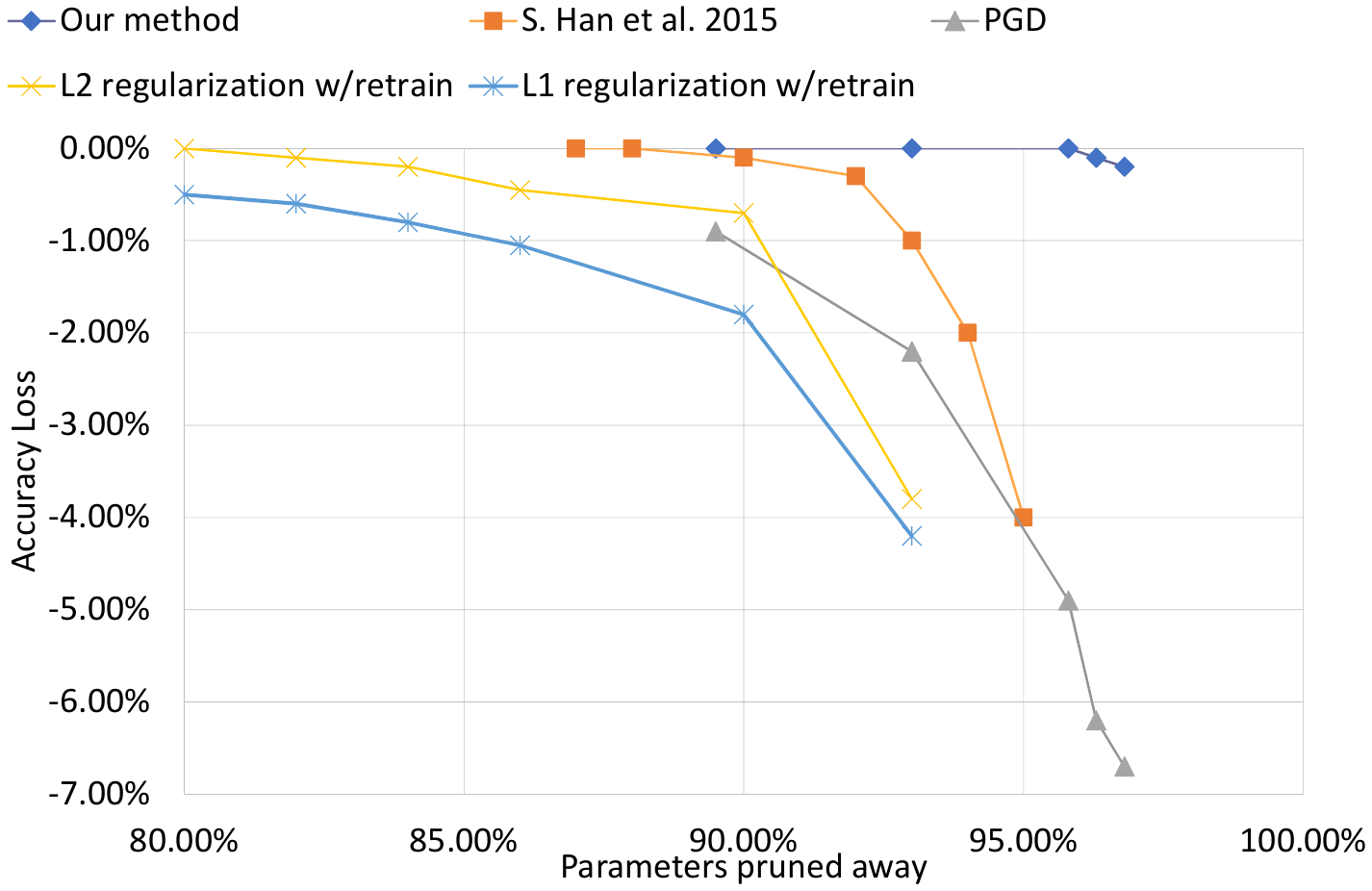}
\caption{Comparison of our method and other methods on non-structured pruning rate of the overall AlexNet model.}
\label{fig:example}
\end{figure}

% \begin{figure*} 
% \centering
% \includegraphics[width=0.75\linewidth]{Drawing1.pdf}
% \caption{ Illustration of 2D weight matrix for GEMM (left) and row-wise and column-wise sparsity (right)}
% \label{fig:example}
% \end{figure*}

\label{sets}
% \textcolor{red}{Yanzhi: I think we need an illustrative figure here.}

%For simplicity of notation, we define ${\bf x}_{b,c,d} := ({\bf{X}})_{:,b,c,d}$.

% \textcolor{red}{Yanzhi: Have you defined $\bf{X}$ before?}

% so on for  
% $
% ({\bf{X}})_{:,:,c,:}
% $
% and
% $
% ({\bf{X}})_{:,:,:,d}

% $.

\begin{table*}[t]
\centering
\caption{Comparison on column sparsity \textbf{without accuracy loss} on AlexNet/CaffeNet model using ImageNet ILSVRC-2012 data set}
\label{my-label}
\begin{threeparttable}
\begin{tabular}{p{3cm}p{2.5cm}p{2.3cm}p{0.7cm}p{0.7cm}p{0.7cm}p{0.7cm}p{0.7cm}p{1.2cm}}
\hline
 Method            & Top1 Acc Loss        & Statistics  & conv1     & conv2 & conv3 & conv4 & conv5 & conv2-5 \\ \hline
 \multirow{2}{*}{SSL \cite{wen2016learning}} & \multirow{2}{*}{0\%} & column sparsity &  0.0\% &    20.9\% &  39.7\%     &  39.7\%     &   24.6\%  & 33.3\%   \\ & & pruning rate &  1.0$\times$ &    1.3$\times$ &  1.7$\times$    &  1.7$\times$     &   1.3$\times$  & 1.5$\times$   \\ \hline
 \multirow{4}{*}{\bf{our method}}                 &     \multirow{4}{*}{0\%}              & column sparsity &  0.0\%    &   70.0\%     &  77.0\%     &    85.0\%   &  81.0\%  & 79.2\%   \\
                  &                  & GPU1$\times$ &1.00$\times$  &2.27$\times$  &3.35$\times$  &3.64$\times$  &1.04$\times$ &2.58$\times$ \\ 
                  &                  & GPU2$\times$ &1.00$\times$  &2.83$\times$  &3.92$\times$  &4.63$\times$  &3.22$\times$ &3.65$\times$  \\ 
& & pruning rate &  1.0$\times$ &    3.3$\times$ &  4.3$\times$    &  6.7$\times$     &   5.3$\times$  & 4.8$\times$   \\                   \hline
\end{tabular}
\begin{tablenotes}
        \footnotesize
        \item[] 
      \end{tablenotes}
    \end{threeparttable}
\end{table*}

\begin{table*}[t]
\centering
\caption{Comparison on column and row sparsity \textbf{with less than 2\% accuracy loss} on AlexNet/CaffeNet using ImageNet ILSVRC-2012}
\label{my-label}
\begin{threeparttable}
\begin{tabular}{p{4cm}p{2.5cm}p{2.3cm}p{0.7cm}p{0.7cm}p{0.7cm}p{0.7cm}p{0.7cm}p{1.2cm}}
\hline
 Method            & Top1 Acc Loss        & Statistics     & conv1          & conv2                & conv3                & conv4                & conv5    & conv2-5            \\ \hline
 \multirow{3}{*}{SSL \cite{wen2016learning}} & \multirow{3}{*}{2.0\%} & column sparsity     &     0.0\%    &     63.2\%                 &   76.9\%                   &            84.7\%          &       80.7\%   &    \multirow{2}{*}{84.4\%\tnote{§}}       \\
                &                   & row sparsity &  9.4\%             &         12.9\%             &      40.6\%                &                     46.9\% &         0.0\%             \\
                  &                  & pruning rate &1.1$\times$  &3.2$\times$  &7.7$\times$  &12.3$\times$  &5.2$\times$ & 6.4$\times$ \\
%                   &                   &       CPU$\times$    &         1.05    &         3.37             &     6.27                 &      9.73                &         4.93             \\
%                   &                   &      GPU$\times$     &         1.00    &         2.37             &       4.94               &     4.03                 &      3.05                \\ 
                  \hline
 \multirow{6}{*}{\bf{our method}} & \multirow{6}{*}{0.7\%} & column sparsity         &     0.0\%    &     63.9\%                 &   78.1\%                   &            87.0\%          &       84.9\% & \multirow{2}{*}{86.3\%\tnote{§}} \\
            &                   & row sparsity            &  9.4\%             &         12.9\%             &      40.6\%                &                     46.9\% &         0.0\%   \\
                  &                  & CPU$\times$ &1.05$\times$  &2.82$\times$  &6.63$\times$  &10.16$\times$  &5.00$\times$ & 6.16$\times$ \\
                  &                  & GPU1$\times$ &1.00$\times$  &1.28$\times$  &4.31$\times$  &1.75$\times$  &1.52$\times$ & 2.29$\times$ \\ 
                  &                  & GPU2$\times$ &1.00$\times$  &2.34$\times$  &6.85$\times$  &6.99$\times$  &4.15$\times$ & 5.13$\times$ \\
                  &                  & pruning rate &1.1$\times$  &3.1$\times$  &7.3$\times$  &14.5$\times$  &6.6$\times$ & 7.3$\times$ \\
                  
 \hline
 \multirow{6}{*}{\bf{our method}} & \multirow{6}{*}{2.0\%} & column sparsity         &     0.0\%    &     87.5\%                 &   90.0\%                   &            90.5\%          &       90.7\% & \multirow{2}{*}{93.7\%\tnote{§}} \\
                &                   & row sparsity            &  9.4\%             &         12.9\%             &      40.6\%                &                     46.9\% &         0.0\%   \\
                  &                  & CPU$\times$ &1.05$\times$  &8.00$\times$  &14.68$\times$  &14.22$\times$  &7.71$\times$  & 11.93$\times$\\
                  &                  & GPU1$\times$ &1.00$\times$  &2.39$\times$  &5.34$\times$  &1.92$\times$  &2.04$\times$  & 3.15$\times$\\ 
                  &                  & GPU2$\times$ &1.00$\times$  &4.92$\times$  &12.55$\times$  &8.39$\times$  &6.02$\times$  & 8.52$\times$\\ 
                  &                  & pruning rate &1.1$\times$  &9.2$\times$  &16.8$\times$  &19.8$\times$  &8.4$\times$ & 15.0$\times$ \\                  \hline
\end{tabular}
\begin{tablenotes}
        \footnotesize
        \item[§] total sparsity accounting for both column and row sparsities.
      \end{tablenotes}
    \end{threeparttable}
\end{table*}

\section{Experiment Results and Discussions}

In this section, we evaluate the proposed StructADMM framework. We first compare on the AlexNet model on ImageNet ILSVRC-2012 data set for a fair comparison with the most related prior work SSL \cite{wen2016learning} on structured pruning.
However, we use a higher baseline AlexNet accuracy of 60.0\% top-1 and 82.4\% top-5 in PyTorch to reflect recent training advances in AlexNet, while the baseline accuracy in SSL is 57.4\% top-1 and 80.3\% top-5. We will conduct a fair comparison because \underline{we focus on the relative accuracy} with our baseline instead of the absolute accuracy (which will of course outperform prior work).  

We directly train the AlexNet while performing structured pruning without using a prior pre-trained AlexNet model, thanks to the compatibility with DNN training of the StructADMM framework.
We set the sparsity by tuning the parameters in the constraint of the optimization problem, which is discussed in Section 4. We initialize these hyperparameters from those explicitly reported in prior work (e.g., non-structured weight pruning \cite{han2015}, SSL \cite{wen2016learning}, etc.). We tune these hyperparameters (increase sparsity level) to find the highest pruning rate we can achieve without accuracy loss by using the bisection method.
Besides the AlexNet model, we also perform evaluation on ResNet for ImageNet data set.

More extensive experiments have been conducted on UCF-101 data set \cite{DBLP:journals/corr/abs-1212-0402} for activity detection, and MNIST and CIFAR-10 \cite{krizhevsky2009learning}. These results, along with convergence analysis of StructADMM, are shown in the supplementary materials link \textcolor{blue}{\url{http://bit.ly/2IBiNBk}}.
We also provide fair comparison with PGD and prior work on filter/channel pruning.

 In the experiments we aim to answer two questions. First, what will be the performance enhancement of the proposed StructADMM framework compared with the prior work on weight pruning? Second, what are the pros and cons for different types of structured sparsity and also non-structured sparsity? To answer the second question, we have conducted extensive speedup testings on two GPU nodes, the high-performance NVIDIA GeForce GTX 1080Ti and the low-power NVIDIA Jetson TX2 (targeting embedded applications), as well as the Intel I7-6700K Quad-Core CPU.

The training of sparse DNN models is performed in PyTorch using NVIDIA 1080Ti and Tesla P100 GPUs. The comparisons on GEMM computation efficiency and acceleration use batch size 1, which is typical for inference \cite{han2015,wen2016learning,wen2017coordinating}. The baseline models and structured sparse models use cuBLAS on GPU and Intel Math Kernel Library (MKL) on CPU. The non-structured sparse models use cuSPARSE library on GPU.

\subsection{Comparison Results on Structured Sparsity}

First, we compare our method with the two configurations of the SSL method \cite{wen2016learning} on AlexNet/CaffeNet. The first has no accuracy degradation (Top-1 error 42.53\%) and average sparsity of 33.3\% on conv2-conv5. We note that the 1st CONV layer of AlexNet/CaffeNet is very small with only 35K weights compared with 2.3M in conv2-conv5, and is often not the optimization focus \cite{wen2016learning,wen2017coordinating}. The second has around 2\% accuracy degradation (Top-1 error 44.66\%) with total sparsity of 84.4\% on conv2-conv5.

Table 1 shows the comparison of our method with the first configuration of SSL.
We generate configuration with no accuracy degradation compared with original model (the original model of our work is higher than that in SSL).
We can achieve a much higher degree of sparsity of 79.2\% on conv2-conv5. This corresponds to 4.8$\times$ weight pruning rate, which is significantly higher compared with 1.5$\times$ pruning in conv2-conv5 in \cite{wen2016learning}. 
%
%Makan comment: I am NOT comfortable with the following sentence:
%These improvements are because ADAM-ADMM involves a systematic way of introducing regularization in DNNs, which should perform better compared with prior work which is based on fixed regularization only.

\begin{table}[t]
\centering
\caption{Structured pruning results on  ResNet-50 for ImageNet}
\begin{tabular}{p{1.9cm}p{5.6cm}}
\hline
Method &    Column prune rate / Top 1 accuracy loss  \\ \hline
Baseline & 1$\times$ / 0\% \\ \hline
\bf{Our method} & 2$\times$ / 0\% \\ \hline
\bf{Our method} & 3$\times$ / 0.9\% \\ \hline
\end{tabular}
\end{table}

\begin{table*}[t]
\centering
\caption{Comparison of our method and other methods on non-structured sparsity on convolutional layers \textbf{without accuracy loss} on AlexNet/CaffeNet model using ImageNet ILSVRC-2012 dataset}
\label{my-label}
\begin{threeparttable}
\begin{tabular}{p{4cm}p{2.5cm}p{1cm}p{0.6cm}p{0.6cm}p{0.6cm}p{0.6cm}p{0.6cm}p{1.15cm}p{1.15cm}}
\hline
 Method            & Top1 acc loss        & Statistics & conv1& conv2 & conv3 & conv4 & conv5 & conv1-5\tnote{§} & conv2-5\tnote{§} \\ \hline
 weight pruning \cite{han2015}                 & 0\%               & sparsity &  16.0\% &  62.0\%     &   65.0\%    &   63.0\%    &  63.0\% & 2.7$\times$ & 2.7$\times$    \\
 dynamic surgery \cite{guo2016dynamic}                 & 0.3\%               & sparsity &  46.2\% &  59.4\%     &   71.0\%    &   67.7\%    &  67.5\% & 3.1$\times$ & 3.1$\times$    \\
 NeST \cite{dai2017}                 & 0\%               & sparsity &  51.1\% &  65.2\%     &   81.5\%    &   61.9\%    &  59.3\% & 3.2$\times$& 3.3$\times$    \\
 fine-grained pruning \cite{mao2017exploring}                 & $-$0.1\% (Top5)               & sparsity &  17.0\% &  74.0\%     &   77.0\%    &   77.0\%    &  77.0\% & 4.2$\times$ & 4.3$\times$    \\
 $l_1$ \cite{wen2016learning} & $-$0.2\% & sparsity  &  14.7\% &  76.2\%     &  85.3\%     &  81.5\%     &  76.3\% & 5.0$\times$ & 5.9$\times$     \\ \hline
 
 \bf{our method}                  &    $-$0.2\%               & sparsity &  0.0\% &  93.0\%     &   94.3\%    &   93.6\%    &  93.5\% & 13.1$\times$ & 16.1$\times$    \\ \hline
\end{tabular}
\begin{tablenotes}
        \footnotesize
        \item[§] Average pruning rate of corresponding layers
      \end{tablenotes}
    \end{threeparttable}
\end{table*}

\begin{table*}[t]
\centering
\caption{Comparison on non-structured sparsity \textbf{within 2\% accuracy loss} on AlexNet/CaffeNet using ImageNet ILSVRC-2012}
\label{my-label}
\begin{threeparttable}
\begin{tabular}{p{3cm}p{2.5cm}p{1cm}p{0.6cm}p{0.6cm}p{0.6cm}p{0.6cm}p{0.7cm}p{1.2cm}p{1.2cm}}
\hline
 Method            & Top1 acc loss        & Statistics & conv1 & conv2 & conv3 & conv4 & conv5 & conv1-5\tnote{§} & conv2-5\tnote{§} \\ \hline 
 $l_1$ \cite{wen2016learning} & 2.0\% & sparsity  &  67.6\% & 92.4\%     &  97.2\%     &  96.6\%     &  94.3\% & 21.8$\times$ & 24.0$\times$ \\ 
 \bf{our method}                  &    0.8\%               & sparsity & 67.6\%  &  92.4\%     &  97.2\%     &  96.6\%     &   94.3\% & 21.8$\times$ & 24.0$\times$   \\

 \bf{our method}                  &    1.95\%               & sparsity & 0.0\%  &  97.0\%     &  98.0\%     &  97.5\%     &   97.0\% & 25.5$\times$ & 40.5$\times$ \\ \hline
\end{tabular}
\begin{tablenotes}
        \footnotesize
        \item[§] Average pruning rate of corresponding layers
      \end{tablenotes}
    \end{threeparttable}
\end{table*}

\begin{table}[t]
\centering
\caption{Comparisons of non-structured weight pruning rate on the whole DNN models for AlexNet and VGGNet}
\begin{tabular}{p{3.4cm}p{2cm}p{2cm}}
\hline
Method &   \small{AlexNet pruning rate / Top 1 acc loss}  & \small{VGG-16 pruning rate / Top 1 acc loss}  \\ \hline
Weight Pruning \cite{han2015}   &  9.0$\times$ / 0\% & 13.0$\times$ / 0.1\% \\ \hline
Optimal Brain Surg. \cite{dong2017learning}  & 9.1$\times$ / 0.3\% & 13.3$\times$ / 0.8\% \\ \hline
Low Rank \& Sparse \cite{yu2017compressing} &  10.0$\times$ / -0.1\% & 15.0$\times$ / 0\% \\ \hline
Dynamic Surgery \cite{guo2016dynamic}  &  17.7$\times$ / 0.3\% & - \\ \hline
\bf{Our method}   & 24.0$\times$ / 0\% & 30.0$\times$ / 0\% \\ \hline

\end{tabular}
\end{table}

% \begin{table*}[t]
% \centering
% \caption{Comparisons of non-structured weight pruning ratio on the whole DNN models for different methods}
% \begin{tabular}{p{7.5cm}p{2.5cm}p{2.5cm}p{2.5cm}}
% \hline
% Method & LeNet-5 pruning ratio / accuracy  &   AlexNet pruning ratio / top 5 accuracy  & VGG-16 pruning ratio / top 5 accuracy  \\ \hline
% Weight Pruning \cite{han2015} &  12.0$\times$ / 99.2\% &  9.0$\times$ / 80.2\% & 13.0$\times$ / 89.1\% \\ \hline
% Net Trim \cite{aghasi2017net} &  45.7$\times$ / 98.7\% &  - & - \\ \hline
% Optimal Brain Surgery \cite{dong2017learning} & -  & 9.1$\times$ / 80.0\% & 13.3$\times$ / 89.0\% \\ \hline
% Low Rank \& Sparse \cite{yu2017compressing}&  - &  10.0$\times$ / 80.3\% & 15.0$\times$ / 89.1\% \\ \hline
% \bf{Our method}  &  87.0$\times$ / 99.2\% & 24.7$\times$ / 80.1\% & 23.4$\times$ / 88.6\% \\ \hline

% \end{tabular}
% \end{table*}

We test the actual GPU accelerations using two GPUs: GPU1 is NVIDIA 1080Ti and GPU2 is NVIDIA TX2. The acceleration rate is computed with respect to the corresponding layer of the original DNN executing on the same GPU and same setup. One can observe that the average acceleration of conv2-conv5 on 1080Ti is 2.58$\times$, while the average acceleration on TX2 is 3.65$\times$. These results clearly outperform the GPU acceleration of 49\% reported in SSL \cite{wen2016learning} without accuracy loss, as well as the recent work \cite{wen2017coordinating}. The acceleration rate on TX2 is higher than 1080Ti because the latter has a high parallelism degree, which will not be fully utilized when the matrix size GEMM of a CONV layer is significantly reduced.

Table 2 shows the comparison with the second configuration of SSL. 
% Again we provide two configurations from ADAM-ADMM. 
With similar (and slightly higher) sparsity in each layer as SSL, we can achieve less accuracy loss. With the same relative accuracy loss (a moderate accuracy loss within 2\% compared with original DNN), a higher degree of 93.7\% average sparsity of conv2-conv5 is achieved, translating into 15.0$\times$ weight pruning. The actual GPU acceleration results are also high: 3.15$\times$ on 1080Ti and 8.52$\times$ on TX2. One can clearly see that the speedup on 1080Ti saturates because the high parallelism degree cannot be fully exploited. The acceleration on CPU can be higher under this setup, reaching 11.93$\times$ on average on conv2-conv5.

Finally, we demonstrate the structured (column) pruning results on ResNet-50 for ImageNet dataset. As shown in Table 3, we achieve 2$\times$ structured pruning with 0\% relative accuracy loss and 3$\times$ structured pruning with 0.9\% accuracy loss. We only demonstrate our results due to lack of prior work for fair comparison.

\subsection{Non-Structured Sparsity on CONV Layers and Comparison Results on Different Platforms}

The StructADMM framework is applicable to non-structured weight pruning as well.
In this section we demonstrate the non-structured weight pruning results (CONV layer pruning) achieved by the StructADMM framework, as well as comparisons with reference work on non-structured pruning \cite{han2015,dai2017,mao2017exploring,guo2016dynamic}. Without any accuracy loss compared with the original AlexNet model, we achieve 16.1$\times$ weight pruning in conv2-conv5, which are 6.0$\times$, 5.2$\times$, 4.9$\times$, 3.75$\times$, and 2.73$\times$ compared with references \cite{han2015}, \cite{guo2016dynamic}, \cite{dai2017}, \cite{mao2017exploring}, and the $l_1$ regularization method in \cite{wen2016learning}, respectively, which is shown in Table 4. Within 2\% accuracy degradation, we achieve 40.5$\times$ weight pruning rate in conv2-conv5, which is shown in Table 5. 

However, even such high degree of sparsity cannot lead to a good GPU acceleration. With 40.5$\times$ weight pruning rate in conv2-conv5, we can only achieve less than 2$\times$ speedup on NVIDIA TX2 and even speed reduction on 1080Ti (where the speed is calculated when both GPUs exploit sparse matrix multiplication). 
This coincides with the conclusion in \cite{wen2016learning} that the high parallelism degree in GPUs cannot be fully exploited in irregular sparsity patterns even if the state-of-the-art sparse matrix multiplication package is utilized.

\subsection{Comparison Results on Overall Non-Structured Compression Rate}

To further demonstrate the potential of our method, we make a fair comparison of our overall non-structured compression rate with the representative works on AlexNet and VGG-16 on the ImageNet ILSVRC-2012 data set. As shown in Table 6, we achieve significantly higher overall non-structured compression rate on AlexNet and VGG-16 than those works without accuracy loss.

\section{Conclusion}

In this paper we proposed a unified, systematic framework of structured weight pruning for DNNs. It is a unified framework for different types of structured sparsity such as filter-wise, channel-wise, shape-wise sparsity as well for non-structured sparsity. In our experiments, we achieve 2.58$\times$ and 3.65$\times$ measured speedup on two GPUs without accuracy loss. The speedups reach 3.18$\times$ and 8.52$\times$ on GPUs and 10.5$\times$ on CPU when allowing moderate accuracy loss of 2\%. Our pruning rate and speedup clearly outperform prior work. 

% \small
% \bibliographystyle{named}
% \bibliography{ijcai19}

\end{document}